\pdfoutput=1
\documentclass[11pt]{article}

\usepackage[review]{ACL2023}

\usepackage{times}
\usepackage{latexsym}
\usepackage{CJK}
\usepackage{multirow}
\usepackage{microtype}
\usepackage{graphicx}
\usepackage{ulem}
\usepackage{amssymb}
\usepackage{booktabs}
\usepackage{arydshln}
\usepackage[T1]{fontenc}
\usepackage[utf8]{inputenc}

\usepackage{microtype}

\usepackage{inconsolata}

\title{CSED: A Chinese Semantic Error Diagnosis Corpus}

\author{Bo Sun\textsuperscript{1}\thanks{\quad indicates equal contribution}, Baoxin Wang\textsuperscript{1,2*}, Yixuan Wang\textsuperscript{1}, Wanxiang Che\textsuperscript{1}\thanks{\quad Corresponding Author: W.Che (car@ir.hit.edu.cn)}, Dayong Wu\textsuperscript{2}, Shijin Wang\textsuperscript{2}, Ting Liu\textsuperscript{1}\\
\textsuperscript{1}Research Center for SCIR, Harbin Institute of Technology, Harbin, China\\
\textsuperscript{2}State Key Laboratory of Cognitive Intelligence, iFLYTEK Research, China\\
\texttt{\{bsun, yixuanwang, car, tliu\}@ir.hit.edu.cn}\\
\texttt{\{bxwang2, dywu2, sjwang3\}@iflytek.com}
}

\begin{document}
\maketitle
\begin{abstract}
% We introduce the corpus of Chinese Semantic Error Diagnosis (CSED), a new dataset to fill the gap in the field of text error correction. 

%before
% Recently, a lot of Chinese text error correction work has focused on Chinese Spelling Check (CSC) and Chinese Grammatical Error Diagnosis (CGED). In contrast, there has been little attention paid to the complicated problem of Chinese Semantic Error Diagnosis (CSED), which is a lack of relevant datasets. The study of semantic errors is significant because of the prevalence of semantic errors in everyday life, particularly among native speakers. To investigate this, we build and release the CSED corpus, which includes two datasets. The one is for the task of Chinese Semantic Error Recognition (CSER). The other is for the task of Chinese Semantic Error Correction (CSEC). Our experiments show that the CSED corpus is not only of high quality but also extremely difficult, as evidenced by the fact that even humans receive a low score. We also find that even powerful pre-trained models do not perform well on this corpus. To this end, we propose syntax-aware approaches to improve model performance. Experiments demonstrate that the syntax-aware approach is helpful under the CSED corpus.

%grammaly
Recently, much Chinese text error correction work has focused on Chinese Spelling Check (CSC) and Chinese Grammatical Error Diagnosis (CGED). In contrast, little attention has been paid to the complicated problem of Chinese Semantic Error Diagnosis (CSED), which lacks relevant datasets. The study of semantic errors is important because they are very common and may lead to syntactic irregularities or even problems of comprehension. To investigate this, we build the CSED corpus, which includes two datasets. The one is for the CSED-Recognition (CSED-R) task. The other is for the CSED-Correction (CSED-C) task. Our annotation guarantees high-quality data through quality assurance mechanisms. Our experiments show that powerful pre-trained models perform poorly on this corpus. We also find that the CSED task is challenging, as evidenced by the fact that even humans receive a low score. This paper proposes syntax-aware models to specifically adapt to the CSED task. The experimental results show that the introduction of the syntax-aware approach is meaningful.

% At the same time, the corresponding datasets are publicly available, such as SIGHAN (dataset for CSC) and NLPCC18 (dataset for CGED).   Due to the complexity of semantic errors, there is no relevant published dataset.   We provide a series of syntax-aware pre-training methods for both CSER and CSEC task. This paper does not attempt to design a state-of-the-art method for CSED but provides the CSED corpus and investigates a more efficient way to make use of syntax. Our experiments show that our method has some improvement in the CSED corpus.

\end{abstract}

\section{Introduction}

Chinese text error correction is widely studied and can be applied in education, journalism, publishing, and other fields. Previous research concentrates more on Chinese Spelling Check (CSC) \cite{jiang2012rule} and Chinese Grammatical Errors Diagnosis (CGED) \cite{lee2015overview}. Meanwhile, the corresponding datasets are publicly available, such as SIGHAN \cite{wu2013chinese,tseng2015introduction} and CGED \cite{rao2020overview}. Conversely, semantic errors are difficult to identify and have not yet attracted the attention of researchers, and there is a lack of relevant datasets. We list the error types for the existing datasets as shown in Table \ref{compare_dataset}. Although some datasets, such as CTC \cite{wang2022cctc} and MuCGEC \cite{zhang-etal-2022-mucgec}, contain semantic errors, they all contain only a small number of semantic errors, which are rare and incomplete. Hence, there is a lack of a CSED corpus containing a rich and comprehensive set of semantic error types. Semantic errors often appear in the Chinese junior or senior high school examination to investigate students' understanding of syntax, semantics, and pragmatics. Semantic errors are also common in everyday life and even problematic for native speakers, leading to syntactic irregularities or even problems of comprehension.  As a result, studying semantic errors is required and essential.

\begin{table}[t]
\centering
\scalebox{0.7}{
\begin{tabular}{cccccc}
\toprule
Major types               & Minor types                  & CGED   & MuCGEC  & CTC    & CSED\\
\hline
           
Spelling                  & -                            & $\checkmark$ & $\checkmark$        & $\checkmark$       & $\times$   \\ \hline
Grammar                   & -                            & $\checkmark$ & $\checkmark$        & $\checkmark$       & $\times$   \\ \hline
\multirow{7}{*}{Semantic} & Word Order                   & $\times$ & $\times$        & $\times$       & $\checkmark$   \\
                          & Missing                     & $\times$  & $\times$        & $\times$       & $\checkmark$   \\
                          & Collocation                  & $\times$ & $\times$        & $\times$       & $\checkmark$   \\
                          & Redundant                    & $\times$ & $\times$        & $\checkmark$       & $\checkmark$   \\
                          & Confusion                    & $\times$ & $\times$        & $\checkmark$       & $\checkmark$   \\
                          & Fuzziness                    & $\times$ & $\checkmark$        & $\times$       & $\checkmark$   \\
                          & Illogic                      & $\times$ & $\checkmark$        & $\times$       & $\checkmark$   \\
\bottomrule
\end{tabular}}
\caption{Comparison of CSED corpus and other datasets.} 
\label{compare_dataset}
\end{table}

\begin{table*}[t]
\centering
\renewcommand\arraystretch{1.2}
\scalebox{0.8}{
\begin{tabular}{lll}
\toprule
Task & Error Type &Sentence \\ \hline
\multirow{2}{*}{CSC}  & \multirow{2}{*}{Spelling Errors} &\begin{CJK*}{UTF8}{gbsn}个人\sout{触须 \textit{<chu xu>}}(储蓄 \textit{<chu xu>})卡存款也有利息吗\end{CJK*}\\
& &  Is there interest on personal debit card deposits \\ \hline
\multirow{8}{*}{CGED} & \multirow{2}{*}{Word Order} & \begin{CJK*}{UTF8}{gbsn}(应该)采取几种方法\sout{应该}帮助他们。\end{CJK*} \\ 
& &(We should) take several methods \sout{should} to help them. \\\cline{2-3}
&\multirow{2}{*}{Missing}&\begin{CJK*}{UTF8}{gbsn}任何婴儿(的)心都是白纸似的清白。\end{CJK*}\\ 
& &The heart (of) any infant is as clear as white paper\\\cline{2-3}
&\multirow{2}{*}{Redundant}&\begin{CJK*}{UTF8}{gbsn}流行歌曲告诉我们现在的\sout{我们的}心理状态。\end{CJK*}\\ 
& &Pop songs tell us about our current \sout{our} state of mind.\\\cline{2-3}
&\multirow{2}{*}{Word Selection}&\begin{CJK*}{UTF8}{gbsn}我晚上写\sout{做 \textit{<zuo>}}(作 \textit{<zuo>})业。\end{CJK*}\\ 
& &I do my homework at night.\\\hline
\multirow{14}{*}{CSED} & \multirow{2}{*}{Word Order} & \begin{CJK*}{UTF8}{gbsn}全厂职工\sout{讨论并听取}(听取并讨论)了报告\end{CJK*}\\
& & The whole staff \sout{discuss and listen}(listen and discuss) the report\\\cline{2-3}
& \multirow{2}{*}{Missing} & \begin{CJK*}{UTF8}{gbsn}这篇报告列举了大量事实，控诉了人类破坏自然，滥杀动物(的行为)。\end{CJK*}\\
& &This report cites many facts and accuses (behaviors of) destroying nature and killing animals. \\\cline{2-3}
& \multirow{2}{*}{Collocation} & \begin{CJK*}{UTF8}{gbsn}我国的汽车产量已经超过法国(，我国)成为全球第四大汽车生产国\end{CJK*}\\
& &Our car production surpassed France (and China) became the world's fourth largest car producer.\\\cline{2-3}
& \multirow{2}{*}{Redundant} & \begin{CJK*}{UTF8}{gbsn}奥斯维辛有\sout{将近}12000余名居民\end{CJK*}\\
& &Auschwitz has \sout{almost} more than 12,000 inhabitants\\\cline{2-3}
& \multirow{2}{*}{Confusion} & \begin{CJK*}{UTF8}{gbsn}由于资金不足\sout{的限制}，学校停止修建图书馆\end{CJK*}\\
& &Due to a lack of funding \sout{constraints}, the school stopped building the library. \\\cline{2-3}
& \multirow{2}{*}{Fuzziness} & \begin{CJK*}{UTF8}{gbsn}山上的水宝贵，我们把它留给\sout{晚上来}(上来晚)的人喝\end{CJK*}\\
& &The water is precious, we leave it to people who come to drink \sout{at night} (late) \\\cline{2-3}
& \multirow{2}{*}{Illogic} & \begin{CJK*}{UTF8}{gbsn}一只鸟在空中\sout{一动不动地}盘旋。\end{CJK*}\\
& &A bird hovers \sout{motionlessly} in the air.\\
\bottomrule
\end{tabular}}
\caption{Examples of different tasks. <*>: Pinyin of Chinese characters.} 
\label{dataset}
\end{table*}
%before
% Unlike spelling and grammatical errors, semantic errors focus on more complex syntax and semantics, leading the sentences with semantic errors to be relatively fluent and even difficult for humans to recognize. 
%grammar
Unlike spelling and grammatical errors, semantic errors focus on more complex syntax and semantics, making sentences with semantic errors relatively fluent and even difficult for humans to recognize. Table \ref{dataset} shows examples of text errors for various tasks and error types. As shown in Table \ref{dataset}, the error type in the CSED task is word order because \begin{CJK}{UTF8}{gbsn}``听取''\end{CJK} (listen) should be placed before \begin{CJK}{UTF8}{gbsn}``讨论''\end{CJK} (discuss) due to the time sequence. In contrast, grammatical errors often lead to incoherent sentences, making it easier for humans to recognize them. For example, in Table \ref{dataset}, the CGED's word order problem is clear, and it causes the entire sentence to be incoherent, which is different from the CSED's word order issue. Semantic errors are a more complex class of text errors that focus more on the syntax and inherent semantics of the entire sentence. The complexity of semantic errors makes the construction of the CSED corpus extremely difficult, which leads to a paucity of data in the CSED task.

To fill the gap in the field of semantic error correction, we build and release the corpus of CSED with two datasets: the CSED-Recognition (CSED-R) dataset and the CSED-Correction (CSED-C) dataset. The CSED-R task is a binary classification task to judge whether a sentence contains semantic errors. The CSED-R dataset, with a total of 49,408 sentences, is produced by multiple-choice questions to determine if they contain semantic errors. The CSED-C task is a natural language generation task that translates incorrect semantic sentences into correct ones. The CSED-C model needs to receive a sentence and output the corrected sentence without semantic errors. The CSED-C dataset is produced and checked by professional annotators with a total of 12,652 sentence pairs.

\begin{figure}[t]
\centering
\includegraphics[width=1.0\columnwidth]{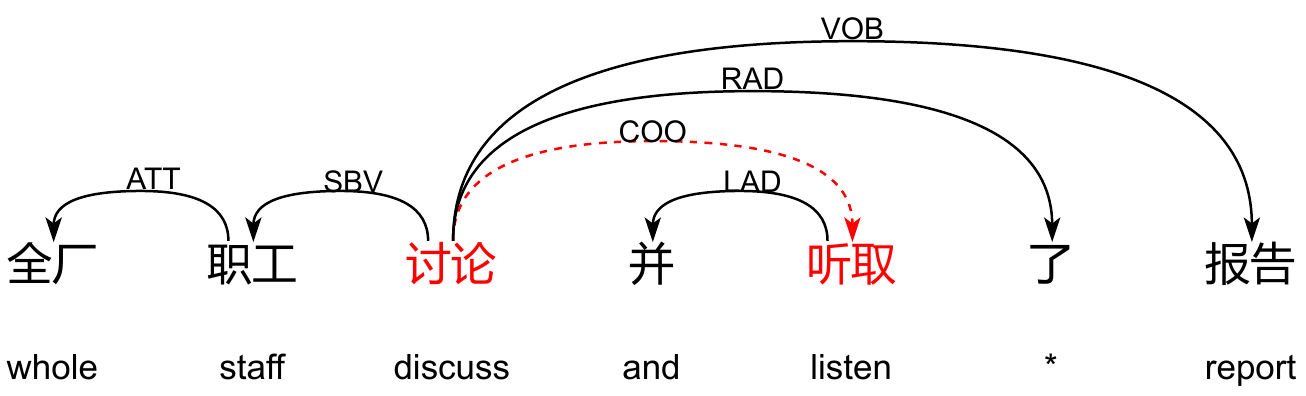} % Reduce the figure size so that it is slightly narrower than the column. Don't use precise values for figure width.This setup will avoid overfull boxes.
\caption{Syntax parsing of incorrect semantic sentences, the incorrect position is marked as red.}
\label{fig-eg}
\end{figure}

Based on the CSED corpus, we propose a series of syntax-aware pre-training approaches for both CSED-R and CSED-C tasks. The reason for the introduction of syntax in the model is that the semantics of a Chinese sentence has a high correlation with syntactic knowledge. For example, as shown in Figure \ref{fig-eg}, the dependency in \begin{CJK}{UTF8}{gbsn}``讨论''\end{CJK} (discuss) and \begin{CJK}{UTF8}{gbsn}``听取''\end{CJK} (listen) is different between the correct and incorrect sentence. We can recognize the semantic error based on the fact that \begin{CJK}{UTF8}{gbsn}``讨论''\end{CJK} (discuss) is the parent node of \begin{CJK}{UTF8}{gbsn}``听取''\end{CJK} (listen) in the incorrect sentence. 
Obviously, it is beneficial for the CSED task to incorporate the syntactic information into the model.

In summary, this paper provides a corpus for Chinese semantic error analysis for the first time. We evaluate this corpus on some representative and capable models. Experimental results show that even the state-of-the-art model does not perform well. To improve the model performance, we propose a syntax-aware approach. The experimental results show that the class of syntax-aware approach improves the performance of CSED tasks. Our main contributions are summarized as follows:
\begin{itemize}
\item We release the CSED corpus, the first corpus for CSED containing two datasets: the CSED-R dataset and the CSED-C dataset.
\item We conduct a detailed analysis for the CSED corpus. First, we elaborate on the differences between the CSED corpus and other existing datasets. Second, we discuss all semantic error types in detail and summarize some characteristics of them.
\item We propose a series of syntax-aware pre-training methods for CSED. Together, our results suggest the need for injecting syntactic information for CSED tasks.

We will release the CSED corpus and codes after the review.
% \item We performed a detailed analysis of the experiment. 
% We analyze what type of semantic error the existing model solves better. Our experiments find that the model performs better on syntax-strongly-related error types and poorly on syntax-weakly-related error types.
% We analyze the distinction of kinds of errors humans make between the CGED and CSED, finding that while CGED is dominated
% by correcting sentence fluency, CSED focuses
% on correcting a broader range of errors that
% appear in the sentence.

\end{itemize}

\section{Related Work}
Text error correction, such as CSC and CGED, has received much attention from researchers. There are already relevant published datasets on the CSC and CGED tasks. SIGHAN Dataset \cite{wu2013chinese,tseng2015introduction} is the earliest spelling error correction dataset. Optical Character Recognition (OCR) Dataset \cite{hong2019faspell} is a pseudo-CSC dataset generated based on OCR technology. Hybrid Dataset \cite{wang2018hybrid} is a pseudo-CSC dataset generated based on OCR and automated speech recognition technology. ECSpell \cite{lv2022general}
is an open multi-domain CSC dataset, including finance, medicine, and other fields. 

For Chinese grammatical errors, The CGED \cite{rao2020overview} series of datasets is oriented to bilingual speakers and contains only grammatical error detection tasks in the early stage and grammatical error correction tasks in the later stage. NLPCC2018 \cite{zhao2018overview} opens grammar error correction evaluation task dataset for bilingual speakers. YACLC \cite{wang2021yaclc} opens CGED dataset for bilingual speakers containing multiple answers. CTC \cite{wang2022cctc} opens CGED dataset for native speakers. MuCGEC \cite{zhang-etal-2022-mucgec} opens CGED dataset for bilingual speakers, containing three domains and multiple answers. Although some datasets, such as those of CTC and MuCGEC, contain a portion of semantic errors, their semantic error types are not comprehensive enough. Therefore, there is a lack of a dataset with a comprehensive set of semantic error types specific to CSED.

% SIGHAN \cite{wu2013chinese,tseng2015introduction}, , ,  (only containing spelling errors) and CGED \cite{rao2020overview}, NLPCC18 \cite{zhao2018overview}, YACLC \cite{wang2021yaclc} (only containing grammatical errors). Conversely, semantic errors are difficult to identify and have not yet attracted the attention of researchers, and there is a lack of relevant datasets. We list the error types for the existing datasets as shown in Table \ref{compare_dataset}. Although some datasets, such as CTC \cite{wang2022cctc} and MuCGEC \cite{zhang-etal-2022-mucgec},

\section{The CSED Corpus}
We introduce the CSED corpus, a set of two datasets: the CSED-R dataset and the CSED-C dataset. The CSED-R task is a binary classification task to judge whether a sentence contains semantic errors. The CSED-R dataset contains pairs (\textit{l}, \textit{s}) where \textit{l} is the label of the sentence \textit{s}, representing whether the sentence contains semantic errors. The CSED-C dataset contains sentence pairs (\textit{s}, \textit{t}). Given a source sentence \textit{s}, the goal of CSED-C is to produce a corrected target sentence \textit{t}.

\subsection{Chinese Semantic Error Recognition}
% before
% In this section, we describe the process of constructing the dataset in detail.In this section, we describe the dataset's construction in detail. We use the web crawler to obtain Chinese multiple-choice questions related to incorrect semantic sentences from junior and senior high school examination online resources. Then we organize these data into a dataset with two labels. One is correct sentences, and the other is incorrect semantic sentences. 
% grammar
In this section, we describe the dataset's construction in detail. First, we use the web crawler to obtain Chinese multiple-choice questions related to incorrect semantic sentences from junior and senior high school examination online resources. Then we organize these data into a dataset with two labels. One is correct sentences, and the other is incorrect semantic sentences. 

We divide these data into train, validation, and test sets. However, some data in the train set is highly similar to the test set, which we call data leakage. To prevent the problem of data leakage, we clean the train set: we delete the data whose text similarity between the validation/test sets and the training set is greater than a fixed threshold $\gamma$. We calculate text similarity by Levenshtein Ratio based on Levenshtein Distance. We select the fixed threshold $\gamma=70\%$ because training data whose text similarity is lower than 70\% is of less similarity compared with the validation and test set. As shown in Appendix \ref{sec:appendix}, we find that the similarity between training and test data is acceptable, and some similar training and test data labels are different. 

Finally, the training dataset contains 45,248 sentences, the validation dataset contains 2,160 sentences, and the test dataset contains 2,000 sentences. More details about our dataset can be seen in Table \ref{CoCLSA}. Since most of the multiple-choice questions we crawl are sentences with semantic errors, there are more sentences with semantic errors in the CSED-R dataset. Therefore, the ratio of sentences with semantic errors are higher in the training set. To ensure reasonableness, we divide the validation and test sets with the same number of correct and incorrect semantic sentences.

\subsection{Chinese Semantic Error Correction}
The CSED-C dataset is completed by human annotation. First, we send 5,000 multiple-choice questions to the annotation company, each with a stem, four options, an answer, and a revision prompt. The annotator's job is to repair semantic errors in each option's sentence by the appropriate revision prompt.

\begin{table}[t]
\centering
\scalebox{0.8}{
\begin{tabular}{c|rcc}
\toprule
   & \#Line & Avg.Length& Error Ratio  \\ \hline
Train   & 45,248 & 50.4 & 74.6\% \\\hline
Dev     & 2,160  & 52.6 & 50.0\% \\\hline
Test    & 2,000  & 54.5 & 50.0\% \\\bottomrule
\end{tabular}}
\caption{Details of the CSED-R dataset where Error Ratio means the proportion of incorrect semantic sentences in the total data.} 
\label{CoCLSA}
\end{table}

We employ thirty employees to work on the annotations. Before the official annotation, each annotator receives training on labeling to improve the quality of labeling. Any issues they encounter while annotating are discussed directly between the annotators and the project manager. Each annotator's output will be randomly sampled and reviewed; any sample with less than 95\% accuracy will be returned and rechecked. 
% We list other labeling specifics here:
% \begin{itemize}
%     \item The annotators must provide two revision results if the revision prompt offers two alternatives.
%     \item The annotators must rigorously adhere to the revision prompt and refrain from making an unneeded and severe revision to the phrase. In addition, the annotators must note any doubtful modifications to provide feedback to the project manager for confirmation.
%     % before
%     % In addition, the annotators must make a note of any doubtful modifications to provide feedback to the project manager for confirmation.
%     \item The annotators must note that we can discard problematic samples, such as those with blank revision prompts or missing titles, when they come across them.
%     \item Because the examples we provide are all collected from the Internet via crawling techniques, they may contain formatting mistakes, redundant characters, and illegal characters. The annotators must review the original text of each option for such errors and make corrections.

% \end{itemize}

Finally, the training dataset contains 10,652 sentences, the validation dataset contains 1,000 sentences, and the test dataset contains 1,000 sentences. Each sentence has an average of 1.2 corrected sentences. More details about our dataset can be seen in Table \ref{CoC}.

% \begin{table}[t]
% \centering
% \scalebox{0.8}{
% \begin{tabular}{c|cc}
% \toprule
% Error Type   & CGED & CSED \\ \hline
% Word Selection   & 44\% & 0 \\
% Word Order     & 6\%   & 13\%  \\
% Missing    & 28\%   & 19\%   \\
% Redundant     & 22\%   & 8\%  \\
% Collocation     & 0  & 24\%  \\
% Confusion     & 0  & 17\%  \\
% Fuzziness     & 0  & 5\%  \\
% Illogic     & 0  & 14\%  \\\bottomrule
% \end{tabular}}
% \caption{Comparison of CGED and CSED dataset.} 
% \label{compare_cged_csed}
% \end{table}

\subsection{How do CGED and CSED errors differ?}
%  before
% To understand how the errors in the CSED
% task differ from errors in the CGED, we compare
% the types of errors in the CSEC and CGEC dataset.  We summarize an error taxonomy that classifies each error. Examples of each error type are shown in Table \ref{dataset}. We find that even for the same error type, CSEC and CGEC have a different focus. For the same error type, CSEC is more difficult than CGEC.
% grammarly
To understand how the errors in the CSED
task differ from errors in the CGED, we compare
the types of errors in the CSED and CGED datasets. We summarize an error taxonomy that classifies each error. Examples of each error type are shown in Table \ref{dataset}. We find that even for the same error type, CSED and CGED have different focuses. For the same error type, CSED is more difficult than CGED.
\begin{itemize}
    \item [(1)]
    \textbf{Word Selection} is a simple error similar to a spelling error, i.e., a word is inappropriate when it appears in a sentence.
    \item [(2)]
    \textbf{Word Order} pays attention to the word-to-word order inside a sentence. The jumbled order of words in the CGED dataset can cause the entire sentence to read poorly. In the CSED dataset, however, the faults are more cryptic, read smoothly, and difficult to identify.
    \item [(3)]
    \textbf{Missing} refers to the absence of one or more words in a sentence. The CGED is mainly missing auxiliaries and prepositions. The CSED is mainly missing subjects, predicates, or objects.
    \item [(4)]
    \textbf{Redundant} refers to the redundancy of one or more words in a sentence. Redundant words in the CGED are mainly exact repetitions of the above or below, that is, the same word repeated twice. The repetition of CSED is mainly semantic; that is, the two words before and after are different in writing but semantically express the same meaning.
    \item [(5)]
    \textbf{Collocation} considers the collocation relationship between words, including subject-verb collocation, verb-object collocation, conjunction collocation, back-and-forth collocation, etc.
    \item [(6)]
    \textbf{Confusion} is a more complex class of semantic error types. Due to the complexity of the Chinese natural language, it is possible to mix two complete sentences inside one sentence, resulting in sentence confusion.
    \item [(7)]
    \textbf{Fuzziness} means that a sentence has two or more different semantics, which is attributed to the phenomenon of multiple meanings of words in Chinese.
    \item [(8)] 
    \textbf{Illogic} refers to the presence of a sentence that does not match the reasoning of the matter.
\end{itemize}

% We randomly sample 100 errors from CGED and CSED datasets, which we then annotate with error types. Error types in the CGED dataset are already marked. Error types sampled from the CSED dataset are annotated based on the reference analysis of the sentence mentioned in Section 2.2. In Table \ref{compare_cged_csed}, we report the error type ratio in the CGED and CSED. The CGED dataset contains four types of grammatical errors. The CSED dataset includes seven types of semantic errors. 

\begin{table}[t]
\centering
\scalebox{0.8}{
\begin{tabular}{c|rccc}
\toprule
   & \#Line & Avg.Length.S& Avg.Length.T & Avg.Edit  \\ \hline
Train   & 10,652 & 52.2 & 51.8 &4.0\\\hline
Dev     & 1,000  & 51.6 & 51.1 &4.2\\\hline
Test    & 1,000  & 52.1 & 51.5 &4.1\\\bottomrule
\end{tabular}}
\caption{Details of the CSED-C dataset where Avg.Length.S means the average length of the source sentence, and Avg.Length.T means the average length of the target corrected sentence. Avg.Edit means the average of edits.} 
\label{CoC}
\end{table}

\section{Approaches to CSED}

\subsection{CSED-R Models}
% before
% We choose the Transformer encoder as our backbone and view the CSER task as a binary classification task. This section introduces dependency-based syntactic knowledge, including dependency structure and dependency relation. Then we propose two kinds of pre-training tasks based on Dependency Structure and Relation Prediction (DSRP) to let models learn the above syntactic knowledge.

% grammar
We choose the Transformer encoder as our backbone and view the CSED-R task as a binary classification task. First, this section introduces dependency-based syntactic knowledge, including dependency structure and dependency relation. Then we propose two pre-training tasks based on Dependency Structure and Relation Prediction (DSRP) to let models learn the above syntactic knowledge.

% \begin{figure*}[htbp]
% \centering
% \includegraphics[width=1.6\columnwidth]{pre-task.pdf} % Reduce the figure size so that it is slightly narrower than the column. Don't use precise values for figure width.This setup will avoid overfull boxes.
% \caption{Structure of our pre-trained model.}
% \label{pre-task}
% \end{figure*}

% before
% Dependency parsing shows a significant improvement in the field of NLP. In this paper, we use the dependency parser of LTP \cite{che2010ltp} to conduct dependency parsing, which provides a series of Chinese natural language processing tools. In order to better represent dependency-based syntactic knowledge, we raise the notion of syntax tree as  $\mathcal{T}=\{\mathcal{R},\mathcal{N},\mathcal{E}\}$, where $\mathcal{R}$ represents the relationship between two nodes, $\mathcal{N},\mathcal{E}$ represents node and edge set.
% grammar
Dependency parsing shows a significant improvement in the field of NLP. In this paper, we use the dependency parser of LTP \cite{che2010ltp} to conduct dependency parsing, which provides a series of Chinese natural language processing tools. Furthermore, to
 better represent dependency-based syntactic knowledge, we raise the notion of syntax tree as  $\mathcal{T}=\{\mathcal{R},\mathcal{N},\mathcal{E}\}$, where $\mathcal{R}$ represents the relationship between two nodes, $\mathcal{N},\mathcal{E}$ represents node and edge set.
\vspace{1ex}

\noindent\textbf{Dependency Structure}\quad Dependency structure considers the directionality of dependency: who is the parent node of two words. From the structure of the syntax tree, the relationship $\mathcal{R}$ includes parent, child, and others. The following $\mathcal{D}(\mathcal{N}_i,\mathcal{N}_j)$ is denoted as the length between node $\mathcal{N}_i$ and $\mathcal{N}_j$, which is the minimal length from node $\mathcal{N}_i$ along the edge to node $\mathcal{N}_j$. The relationship $\mathcal{R}$ can be expressed as follows: $\mathcal{R}_{ij}=child (parent)$ if $\mathcal{N}_i$ is child (parent) node of $\mathcal{N}_j$ and $\mathcal{D}(\mathcal{N}_i,\mathcal{N}_j)=1$, otherwise $\mathcal{R}_{ij}=others$.
As shown in Figure \ref{fig-eg}, for example, $\mathcal{R}$(\begin{CJK}{UTF8}{gbsn}全厂\end{CJK},\begin{CJK}{UTF8}{gbsn}职工\end{CJK})$=parent$, $\mathcal{R}$(\begin{CJK}{UTF8}{gbsn}职工\end{CJK},\begin{CJK}{UTF8}{gbsn}全厂\end{CJK})$=child$  and $\mathcal{R}$(\begin{CJK}{UTF8}{gbsn}了\end{CJK},\begin{CJK}{UTF8}{gbsn}报告\end{CJK})$=others$. Since there are many relationships except $child$ and $parent$, we classify those relationships that are more than one distance ($\mathcal{D}(\mathcal{N}_i,\mathcal{N}_j)>1$) as $others$. Hence, the relationship $\mathcal{R}_{ij}=others$ contains many types: sibling, grandparent, etc. 

\vspace{1ex}

\noindent\textbf{Dependency Relation}\quad Dependency relation considers the diversity of dependencies, that is, what is the specific dependency relation between two words. Through syntactic dependency parsing, we can find that different words have different dependency relations. As shown in Figure \ref{fig-eg}, for example, $\mathcal{R}$(\begin{CJK}{UTF8}{gbsn}职工\end{CJK},\begin{CJK}{UTF8}{gbsn}全厂\end{CJK})$=ATT$ and $\mathcal{R}$(\begin{CJK}{UTF8}{gbsn}讨论\end{CJK},\begin{CJK}{UTF8}{gbsn}职工\end{CJK})$=SBV$. 

% According to the results of dependency parsing for LTP, we summarize 12 dependency relations as shown in Table \ref{tags}.

\vspace{1ex}

\noindent\textbf{Dependency Prediction Task}\quad We have the following pre-training tasks. The first one is MLM, the same as BERT. Another pre-training task is Dependency Structure and Relation Prediction (DSRP), which is proposed to allow the pre-trained model to learn the syntactic information from dependency parsing. We randomly select some pairs of Chinese words and let the model predict the dependency between them. We use pre-trained models to generate the representation of the last hidden states of the pairs of Chinese words we selected. Since Chinese words consist of multiple tokens, we put these Chinese tokens into a pooling layer with max-pooling. Then we put it into the classifier for classification tasks. In this paper, we select Multilayer Perceptron (MLP) as the classifier consisting of 4 layers. We select Rectified Linear Unit as an activation function in MLP.

According to syntactic knowledge of dependency structure and dependency relation, we have the following pre-training tasks: Dependency Structure Prediction (DSP), Dependency Relation Prediction (DRP), and Dependency Structure and Relation Prediction (DSRP).
\begin{itemize}
    \item DSP: This pre-training task only considers two dependency structures, including $child$ and $parent$. We randomly select some pairs of Chinese words whose dependency structure is either $child$ or $parent$ and let the model predict these dependency structures. The pre-trained models can learn the directionality of the dependency structure in this pre-training task.
    \item DSP$^+$: In this pre-training task, we consider three dependency structures, including $child$, $parent$ and $others$. DSP$^+$ is similar to DSP, but the only difference is that the number of dependent structures considered by the two pre-training tasks is different. This pre-training task considers all the dependency structures and is thus a variant of DSP.
    \item DRP: In this pre-training task, we consider 12 dependency relations. We randomly select some pairs of Chinese words with 12 dependency relations using the dependency parser of LTP. The pre-trained models can learn the diversity of dependency relation in this pre-training task.
    \item DSRP: We combine DSP and DRP for multi-task training. 
    \item DSRP$^+$: We combine DSP$^+$ and DRP for multi-task training. 
\end{itemize}

\begin{table}[t]
\renewcommand\arraystretch{1.1}
\centering
\scalebox{0.7}{
\begin{tabular}{lccc}
\toprule
Model   & $P$& $R$& $F_1$ \\ \hline
\multicolumn{4}{c}{\textit{General Pre-trained Models}}\\\hline
BERT \cite{devlin2018bert}          &$71.5_{\pm1.4}$    &$72.2_{\pm1.2}$    &$71.9_{\pm0.6}$\\
BERT+wwm \cite{cui2019pre}          &$71.1_{\pm0.6}$    &$74.4_{\pm0.3}$    &$72.7_{\pm0.2}$\\
ERNIE1.0 \cite{sun2019ernie}        &$70.4_{\pm0.3}$    &$77.2_{\pm0.2}$    &$73.7_{\pm0.2}$\\
RoBERTa \cite{liu2019roberta}       &$72.9_{\pm0.5}$	&$72.4_{\pm1.6}$    &$72.6_{\pm0.7}$\\
RoBERTa+wwm                         &$72.4_{\pm0.6}$    &$75.0_{\pm1.1}$    &$73.7_{\pm0.3}$\\
MacBERT \cite{cui2020revisiting}    &$72.3_{\pm0.7}$    &$75.3_{\pm1.5}$    &$73.7_{\pm0.4}$\\
\hline
\multicolumn{4}{c}{\textit{RoBERTa Fine-tuning with Syntax-Infused Models}}\\\hline
SLA \cite{li2020improving}          &$72.8_{\pm0.6}$    &$73.0_{\pm1.3}$    &$72.9_{\pm0.6}$\\
Syntax-RoBERTa \cite{bai2021syntax} &$73.3_{\pm0.2}$    &$74.3_{\pm0.4}$    &$73.8_{\pm0.2}$\\
\hline
\multicolumn{4}{c}{\textit{RoBERTa Pre-training with Syntax-related Task}}\\\hline
K-adapter \cite{wang2020k}          &$72.6_{\pm0.8}$	&$73.7_{\pm0.9}$    &$73.2_{\pm0.2}$\\
\hdashline
RoBERTa+DSRP                        &$74.2_{\pm0.5}$	&$74.4_{\pm1.5}$    &$74.3_{\pm0.5}$\\
RoBERTa+DSRP$^+$                    &$73.2_{\pm1.0}$    &$75.8_{\pm2.1}$    &$74.8^\spadesuit_{\pm0.3}$\\

SLA + DSRP                          &$72.1_{\pm1.1}$    &$77.1_{\pm1.7}$    &$74.5_{\pm0.2}$\\
SLA + DSRP$^+$                      &$72.0_{\pm0.6}$    &$76.9_{\pm0.9}$    &$74.4_{\pm0.3}$\\

Syntax-RoBERTa + DSRP               &$73.7_{\pm0.6}$	&$75.9_{\pm1.3}$    &$74.8_{\pm0.4}$\\
Syntax-RoBERTa + DSRP$^+$           &$73.6_{\pm0.8}$    &$76.1_{\pm1.8}$    &$74.8_{\pm0.6}$\\
\hline

MacBERT + DSRP                      &$73.6_{\pm0.6}$    &$75.9_{\pm1.3}$    &$74.7_{\pm0.6}$\\
MacBERT + DSRP$^+$                  &$71.5_{\pm0.9}$	&$78.8_{\pm2.1}$    &$\textbf{74.9}^\spadesuit_{\pm0.6}$\\\hline
Human                               &$72.4_{\pm3.1}$    &$78.6_{\pm8.7}$    &$75.1_{\pm3.7}$\\
\bottomrule
\end{tabular}}
\caption{We report the average score and standard deviation of 3 independent runs with different seeds. For the convenience of understanding, we make the following explanation. DSRP: DSP+DRP, DSRP$^+$: DSP$^+$+DRP, DSP: 2-dependency structure, DSP$^+$: 3-dependency structure, DRP: 12-dependency relation. $\spadesuit$ means our improvement compared with general pre-trained models and Syntax-Infused models is statistically significant with $p < 0.05$ under the t-test.}\label{main_results}
\end{table}

\subsection{CSED-C Models}
We choose mT5 \cite{xue2020mt5} as our backbone and consider the CSED-C task as a machine translation task, i.e., translating sentences containing semantic errors into correct sentences. This section provides a syntax-aware pre-training approach, a pseudo-data construction method to solve the problem of insufficient training data for the CSED-C task.

\vspace{1ex}

\noindent\textbf{Word Order of Adverbial Adjunct and Attribute}\quad In Chinese, the adverbial adjunct should modify the verb, while the attribute should modify the object. Hence, if the adverbial adjunct modifies the object or the attribute modifies the verb, this leads to the word order of adverbial adjunct and attribute. 

\vspace{1ex}

\noindent\textbf{Word Order of Conjunctions
}\quad In Chinese, if the subjects of two clauses are different, the subject should be placed after the conjunction. If the subjects of the two clauses are the same, the subject should be placed before the conjunction. We obtain the subject of the sentence by dependency parsing and destroy the sentence according to the above linguistic rule.

\vspace{1ex}

\noindent\textbf{Missing of Subject or Predicate or Object}\quad We get the subject, predicate, and object of the sentence according to the dependency parsing and delete one randomly. To make the constructed data as close as possible to the actual data, we avoid deleting entities, which would make the meaning of the sentence confusing.

\section{Experiments on the CSED-R dataset}
\subsection{Experimental setup}
We use 1 million Wikipedia data as a pre-training dataset in the pre-training stage. We use LTP as a tool for syntactic parsing.\footnote{\url{http://ltp.ai/}} We take RoBERTa \cite{liu2019roberta} as the base pre-trained model and pre-train for 10 epochs with an effective batch size of 256. We use AdamW optimizer \cite{kingma2014adam,loshchilov2017decoupled} with a learning rate of 2e-5 and weight decay of 0.01. We use a learning rate warmup for 2,500 steps. In the fine-tuning stage, we use the CSED-R dataset as a fine-tuning dataset. We fine-tune the pre-trained models for 4 epochs with an effective batch size of 32. Finally, we report the F1 score of sentences with semantic errors. The implementation of pre-training and fine-tuning is based on HuggingFace's Transformer \cite{wolf2019huggingface}, which consists of 12-layer, 768-hidden, and 12-heads.

\subsection{Results}
Table \ref{main_results} demonstrates the results of different models on the CSED-R task. Overall, our approaches improve general pre-trained models and Syntax-Infused models. Moreover, the improvement of our model compared with the baseline is statistically significant with $p <0.05$ under the t-test.

\textbf{It is useful to introduce syntactic information into the pre-trained model for the CSED-R task.} RoBERTa+DSRP/DSRP$^+$ achieves an improvement of 1.7\%/2.2\% in $F_1$ score compared with RoBERTa. Compared with the strongest baseline MacBERT, RoBERTa+DSRP/DSRP$^+$ has a 0.6\%/1.0\% improvement in the $F_1$ score. This result indicates that our methods outperform general pre-trained models for the CSED-R task.

\begin{figure*}[t]
\centering
\includegraphics[width=1.7\columnwidth]{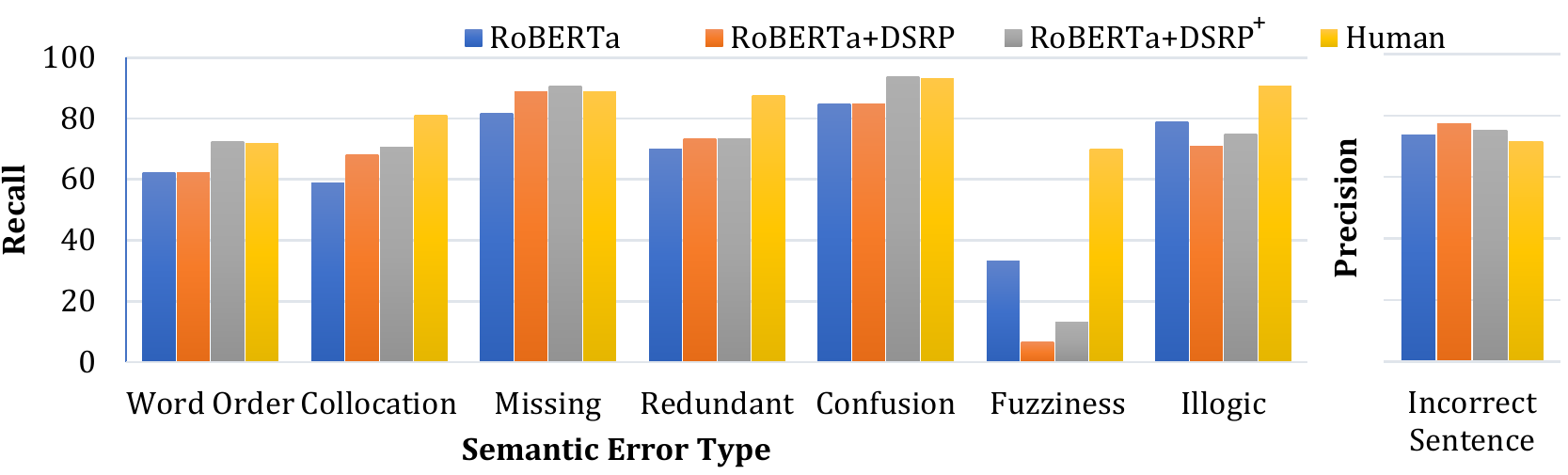} % Reduce the figure size so that it is slightly narrower than the column. Don't use precise values for figure width.This setup will avoid overfull boxes.
\caption{The recognition ability of the models for different types of semantic errors. We report the average score of 3 independent runs with different seeds for models and the average score of 4 people for the human level.}
\label{histogram}
\end{figure*}
% before
% RoBERTa+DSRP/DSRP$^+$ reaches an improvement of 1.1\%/1.6\% in $F_1$ score compared with K-adapter. The result of the K-adapter model is not as good as ours because the syntax-related pre-training task in K-adapter is insufficient. In contrast, our pre-training tasks consider the directionality of dependency structure and the diversity of dependency relation. Hence, our models surpass the K-adapter for the CSER task.
% grammar
RoBERTa+DSRP/DSRP$^+$ reaches an improvement of 1.1\%/1.6\% in $F_1$ score compared with K-adapter. The result of the K-adapter model is not as good as ours because the syntax-related pre-training task in K-adapter is insufficient. In contrast, our pre-training tasks consider the dependency structure's directionality and the dependency relation's diversity. Hence, our models surpass the K-adapter for the CSED-R task.

% before
% We also conduct DSRP and DSRP$^+$ pre-training tasks on the most potent pre-trained model MacBERT. MacBERT+DSRP/DSRP$^+$ achieves an improvement of 1.0\%/1.2\% in $F_1$ score compared with MacBERT. This result indicates that our method is significantly improved even on powerful pre-trained models, which means that it can be used in the various pre-trained models to increase syntax knowledge perception.
% grammar
We also conduct DSRP and DSRP$^+$ pre-training tasks on the most potent pre-trained model MacBERT. MacBERT+DSRP/DSRP$^+$ achieves an improvement of 1.0\%/1.2\% in $F_1$ score compared with MacBERT. This result indicates that our method is significantly improved even on powerful pre-trained models, which can be used in the various pre-trained models to increase syntax knowledge perception.

% before
% \textbf{It is more effective to make use of syntactic information for the CSER task in the pre-training stage rather than in the fine-tuning phase.} RoBERTa+DSRP/DSRP$^+$ achieves better results than SLA in 1.4\%/1.9\% $F_1$ score. RoBERTa+DSRP/DSRP$^+$ gains better results than Syntax-RoBERTa in 0.5\%/1.0\% $F_1$ score. This result reveals that it is more effective for the CSER task to let the model learn syntactic knowledge in the pre-training stage than to inject syntactic knowledge directly.
% grammar
\textbf{It is more effective to use syntactic information for the CSED-R task in the pre-training stage rather than in the fine-tuning phase.} RoBERTa+DSRP/DSRP$^+$ achieves better results than SLA in 1.4\%/1.9\% $F_1$ score. RoBERTa+DSRP/DSRP$^+$ gains better results than Syntax-RoBERTa in 0.5\%/1.0\% $F_1$ score. This result reveals that it is more effective for the CSED-R task to let the model learn syntactic knowledge in the pre-training stage than injecting it directly.

\textbf{Ours can further improve Syntax-Infused models for the CSED-R task.} Comparing to Syntax-RoBERTa, Syntax-RoBERTa+DSRP/DSRP$^+$ brings an improvement of 1.4\%/1.0\% in $F_1$ score. Compared to SLA, SLA+DSRP/DSRP$^+$ obtains an improvement of 1.6\%/1.5\% in the $F_1$ score. The method in Syntax-Infused models and ours based on novel pre-training tasks are two completely different ideas. Syntax-Infused models directly incorporate syntactic information into the model in the fine-tuning stage. In contrast, we design some dependency-related pre-training tasks to let the model learn syntactic information in the pre-training stage. This result demonstrates that our methods enhance Syntax-Infused models by taking our methods in the pre-training stage.

% \begin{table}[t]
% \centering
% \scalebox{0.8}{
% \begin{tabular}{cccc}
%     \toprule
%     Model    & $P$& $R$& $F_1$ \\
%     \midrule
%     RoBERTa &  74.0  & 71.6 & 70.9   \\
%     RoBERTa+DSRP & 75.4 & 73.3 & 72.7       \\
%     RoBERTa+DSRP$^+$ & 75.3 & 73.5 & \textbf{73.0} \\
%     \bottomrule
% \end{tabular}}
% \caption{Experimental results of our models and baseline for CGED dataset.}
% \label{cged:result}
% \end{table}

\subsection{Discussion}
% \textbf{Our methods can also improve the model on the task of CGED.}\quad To explore the ability of our model on the task of CGED, we select the detection level of CGED as our target task. Similar to CSER, the detection level of CGED is a binary classification task that focuses on whether a sentence contains grammatical errors, including redundant, missing, selection, and word order. We construct a dataset of CGED as shown in Table \ref{cged}. We sample an equal number of correct sentences and sentences with grammatical errors from the training sets of CGED 2016 and CGED 2017. Then we remove the duplicate part of the data and divide it into a new training set, validation set, and test set. More details can be seen in Table \ref{cged}. 

% We can see the experimental results in Table \ref{cged:result}. RoBERTa+DSRP/DSRP$^+$ has an improvement of 1.8\%/2.1\% in $F_1$ score compared with RoBERTa. The result demonstrates that our methods also perform well on the CGED task. This advancement is due to the similarity of some types of grammatical and semantic errors. For example, \textit{word order} problems not only appear on the CGER task but also on the CSER task. The recognition of these errors benefits from syntactic knowledge. The difference is that the \textit{word order} errors of CGED are often the disorder between any two words, which will affect the fluency of sentences. However, in the CSER task, it drives semantic changes without affecting sentence fluency. 

% \vspace{1ex}

\noindent\textbf{The syntax-strongly-related error types in the CSED-R dataset can benefit more from syntax.}\quad How is the recognition ability of the model under various types of semantic errors? To figure this out, we randomly sample 200 sentences from our test set, including 100 correct and 100 incorrect sentences. Because CSED-R is a binary classification task, we can only calculate the standard recall score for a specific type of semantic error. In order to comprehensively measure the recognition ability of the model in different error types, we also list the precision score for semantic errors as a reference. If the recall score of a specific semantic error is high and the overall precision score is also high, the model performs well in this semantic error.
We list the result in Figure \ref{histogram}. Compared to our baseline RoBERTa, our methods perform better for some semantic error types, such as \textit{word order}, \textit{collocation}, \textit{missing}, \textit{redundant}, and \textit{confusion}. These error types are strongly related to syntactic information. This result proves that our model does learn practical syntactic knowledge during the pre-training stage. However, our method's recall ability is not as good as the baseline on the semantic error types of \textit{fuzziness} and \textit{illogic}. These errors have little to do with the syntax but more with global semantic information. That is to say, letting the model learn syntactic information cannot solve this kind of problem but reduces the recall ability of this type of error because the pre-training task concentrates on syntax.

However, humans get lower recall scores in \textit{word order} and \textit{fuzziness} error types. This may be because people tend to pay less attention to word order when speaking in daily life. Some inversions of word order do not affect human understanding of the sentences, so humans are not so ``strict" on word order issues. Furthermore, \textit{fuzziness} is relatively obscure to humans, and these sentences often appear complete. Hence, humans are weak in the identification of such errors. In addition, humans have the lowest precision score compared to models.

\vspace{1ex}

\noindent\textbf{The CSED-R task is challenging for even humans.}\quad To explore college humans' performance on the CSED-R task, we hired four students from a top-ranking university and paid remuneration, including two undergraduate students, one graduate student, and one doctoral student. In order to ensure the quality of the labeling results, we let these students label the data independently without outside help. The results show that our model is closest to human performance and slightly lower than humans in the F1 score. This proves that the CSED-R task is challenging for the model and needs further improvement. Human performance on the CSED-R task can be seen in Table \ref{main_results}.

\section{Experiments on the CSED-C dataset}
\subsection{Experimental setup}
We use 1 million pseudo-data conducted by the rule mentioned in Section 3.2. We take mT5 \cite{xue2020mt5} as our backbone and pre-train for 20 epochs with an effective batch size of 128. In the fine-tuning stage, we use the CSED-C dataset as a fine-tuning dataset. We fine-tune the model for 10 epochs with an effective batch size of 32. Inheriting the metric calculation method of previous researchers, we report the F0.5 score using MaxMatch scorer \cite{zhang-etal-2022-mucgec}.

\begin{table}[t]
\centering
\scalebox{0.8}{
\begin{tabular}{ccccc}
    \toprule
    Model &Pseudo-data   & $P$& $R$& $F_{0.5}$ \\
    \midrule
    mT5-small & $\times$& 33.7  & 5.4 & 16.5   \\
    mT5-small &\checkmark& 54.3 & 15.4 & 36.1       \\
    mT5-base& $\times$& 57.0 & 19 & 40.7 \\
    mT5-base &\checkmark& 53.0 & 27.8 & 44.9 \\
    BART-large& $\times$& 53.8 & 38.3 & 49.7 \\
    BART-large& \checkmark& 51.0 & 39.3 & 48.1 \\
    SynGEC$^\spadesuit$& $\times$& 53.0 & 39.5 & 49.6 \\
    Human& $\times$& 52.0 & 41.9 & 49.5 \\
    \bottomrule
\end{tabular}}
\caption{Experimental results of our models and baseline for CSED-C task. $\spadesuit$: the state-of-the-art model of CGED task. Human: average of three people sampling 100 sentences from the test set in the CSED-C dataset.}
\label{csec}
\end{table}

\subsection{Results \& Discussion}
Table \ref{csec} shows that the mT5 model can benefit from our pre-training method via pseudo-data construction. However, the BART \cite{shao2021cpt} model does not improve under the pseudo-data construction method. This is attributed to the relatively high recall of the BART model itself, which is already difficult to improve with high recall with the pseudo-data pre-training approach. On the contrary, the mT5 model itself has a relatively low recall, so the pseudo-data pre-training approach can improve the mT5 model.

\textbf{Can CGED models be directly adapted for CSED-C?} Since CSED is structurally identical to CSED-C, a natural question is whether models which are the state-of-the-art model of CGED can be directly adapted for CSED-C. SynGEC\cite{zhang2022syngec} is an improved model of BART-large using syntax for the CGED task. The results show that even the state-of-the-art model of CGED performs poorly under the CSED-C.

\textbf{How difficult is the CSED-C task?} To quantify how difficult the CSED-C task is, we report the human score in Table \ref{csec}. Three master's degree students are randomly selected to participate in the assessment, given the task of revising a given sentence into a correct one. Participants are required to complete 100 sentences, randomly selected from the test set, within two hours. The results show that the CSED-C task is indeed challenging because humans also score lower on this task.

\section{Conclusion \& Future Work}
We introduce and release the Chinese Semantic Error Diagnosis (CSED) corpus with two datasets to study the CSED-R task and the CSED-C task. In our analysis of CSED data, we show how the errors that humans make differ from those made in CGED. The CSED corpus contains richer semantic error types compared to other existing datasets. We find that various powerful models can not solve this task well. In addition, we report the human score on this task and find that even if humans perform poorly, proving the difficulty of the CSED task.  The experimental results show that the introduction of the syntax-aware approach is meaningful. However, even with the addition of a syntax-aware method, we discover that the model does not perform well on specific error types. Our future study will focus more on external knowledge to improve the model's performance.

\clearpage
\section*{Limitations}
% before
% First, the CSED corpus is mainly for Chinese, although there are semantic errors in other languages, such as English. Second, our dataset is not labeled with the error type of the sentence because it requires some expertise to determine the error type. We will then organize professionals to mark the types of errors in the sentences.
% grammar
First, the CSED corpus is mainly for Chinese, although semantic errors exist in other languages, such as English. Second, our dataset is not labeled with the error type of the sentence because it requires some expertise to determine the error type. We will then organize professionals to mark the types of sentence errors.

% ACL 2023 requires all submissions to have a section titled ``Limitations'', for discussing the limitations of the paper as a complement to the discussion of strengths in the main text. This section should occur after the conclusion, but before the references. It will not count towards the page limit.
% The discussion of limitations is mandatory. Papers without a limitation section will be desk-rejected without review.

% While we are open to different types of limitations, just mentioning that a set of results have been shown for English only probably does not reflect what we expect. 
% Mentioning that the method works mostly for languages with limited morphology, like English, is a much better alternative.
% In addition, limitations such as low scalability to long text, the requirement of large GPU resources, or other things that inspire crucial further investigation are welcome.

\section*{Ethics Statement}
For the data from the CSED-R dataset, the information we collect is through legal channels or from public resources. If it comes from other places, it is also allowed and authorized and will not violate any code of ethics. For the annotated data from the CSED-C dataset, we have paid the annotators. We annotate a total of 5,000 multiple-choice questions at a rate of 2.6 RMB per multiple-choice question. Each additional revision results in an additional payment of 1 RMB. For questions that cannot be modified, we pay 0.5 RMB. We report the full text of instructions given to participants, including e.g., screenshots, disclaimers of any risks to participants or annotators, etc. Annotators have a bachelor's degree and specialize in data annotation.

For the human performance test on CSED tasks, we inform the participants of the purpose of the study in advance and pay the remuneration. They will not disclose or infringe on any privacy during the study. They can stop participating at any time. In short, we abide by all research ethics.

\normalem

\bibliography{acl2023}
\bibliographystyle{acl_natbib}
% \nocite{Ando2005,augenstein-etal-2016-stance,andrew2007scalable,rasooli-tetrault-2015,goodman-etal-2016-noise,harper-2014-learning}
\appendix
\newpage
\section{Appendix}
\label{sec:appendix}
Data leakage means that the data in the training set and the test set are the same or highly similar. This paper uses the Levenshtein ratio as the similarity score between texts. We clean the data of the train set with a similarity score greater than 70\% between the train and test set. Because we find that sentences with a similarity score lower than 70\% can be considered to have no data leakage problem. We enumerate the top-5 sentence pairs with the highest similarity between cleaned train and test sets as shown in Table \ref{data_overlap}. In Case 1-2, the data in the train and test sets are not similar. In Case 3-5, the sentence labels of the train set and the sentence labels of the test set are even different. Therefore, we believe that our test set does not have the problem of data leakage.

\begin{table*}[htbp]
\renewcommand\arraystretch{1.2}
\scalebox{0.9}{
\begin{tabular}{llll}
\toprule
Case              & dataset & Sentence & Label \\ \hline
\multirow{2}{*}{1} & train  & \begin{CJK*}{UTF8}{gbsn}在激烈的市场竞争中，博兰公司所缺乏的，一是创意不佳，二是资金不足。\end{CJK*}& incorrect   \\
& test&\begin{CJK*}{UTF8}{gbsn}在激烈的市场竞争中，很多企业所缺乏的，一是勇气不足，二是谋略不当。\end{CJK*}& incorrect \\ \hline
\multirow{4}{*}{2} & \multirow{2}{*}{train}  & \begin{CJK*}{UTF8}{gbsn}互联网不仅能浏览信息、收发电子邮件，还可以提供网上视频点播和远程\end{CJK*}& \multirow{2}{*}{incorrect}   \\
&&\begin{CJK*}{UTF8}{gbsn}教学等智能化、个性化。\end{CJK*}& \\
& \multirow{2}{*}{test}&\begin{CJK*}{UTF8}{gbsn}宽带网络作为信息社会的主要纽带，它不仅能浏览信息，还可以提供网上\end{CJK*}& \multirow{2}{*}{incorrect} \\
&&\begin{CJK*}{UTF8}{gbsn}视频点播和远程教育等智能化、个性化。\end{CJK*}& \\\hline
\multirow{2}{*}{3} & train  & \begin{CJK*}{UTF8}{gbsn}劳动工资的改革，对某些吃惯“大锅饭”的职工，的确会感到不适应。
\end{CJK*}& incorrect \\
& test&\begin{CJK*}{UTF8}{gbsn}某些吃惯“大锅饭”的职工对劳动工资制度的改革，的确会感到不适应。
\end{CJK*}& correct \\\hline
\multirow{4}{*}{4} & \multirow{2}{*}{train}  & \begin{CJK*}{UTF8}{gbsn}只有充分地对于一个问题的两方面的事实和论点加以叙述和比较，才能得\end{CJK*}& \multirow{2}{*}{incorrect}   \\
&&\begin{CJK*}{UTF8}{gbsn}到良好的结果，但这里不可能这样做。\end{CJK*}& \\
& \multirow{2}{*}{test}&\begin{CJK*}{UTF8}{gbsn}我们只有对一个问题的两方面的事实和论点加以充分地比较和叙述,才能得\end{CJK*}& \multirow{2}{*}{correct} \\
&&\begin{CJK*}{UTF8}{gbsn}到良好的结果。\end{CJK*}& \\\hline
\multirow{4}{*}{5} & \multirow{2}{*}{train}  & \begin{CJK*}{UTF8}{gbsn}随着求职竞争的加剧，招聘企业不仅注重学历、文凭等硬指标，也日益看\end{CJK*}& \multirow{2}{*}{correct}   \\
&&\begin{CJK*}{UTF8}{gbsn}重求职者的工作热情、责任心与沟通能力等“软指标”。\end{CJK*}& \\
& \multirow{2}{*}{test}&\begin{CJK*}{UTF8}{gbsn}随着竞争的加剧，招聘企业不仅注重求职者的工作热情、责任心与沟通能\end{CJK*}& \multirow{2}{*}{incorrect} \\
&&\begin{CJK*}{UTF8}{gbsn}力等“软指标”，也日益看重求职者的学历、文凭等硬指标。\end{CJK*}&  \\\bottomrule
\end{tabular}
}
\caption{Top-5 sentence pairs with the highest similarity between train and test sets.} \label{data_overlap}
\end{table*}

\end{document}